\documentclass[10pt,twocolumn,letterpaper]{article}

\usepackage[pagenumbers]{wacv} 

\usepackage[accsupp]{axessibility}
\usepackage{graphicx}
\usepackage{amsmath}
\usepackage{amssymb}
\usepackage{booktabs}

\usepackage{bbm}
\usepackage{mathtools}

\usepackage{algorithm}
\usepackage{algpseudocode}
\usepackage{makecell}
\usepackage{soul} 

\usepackage{placeins} 
\usepackage{titletoc} 

\usepackage{tikz}
\usetikzlibrary{shapes}

\usepackage[font=small,labelfont=bf]{caption}

\usepackage{pgf}

\newcommand{\R}{\mathbb{R}}
\newcommand{\SN}{\mathcal{N}}
\newcommand{\SL}{\mathcal{L}}
\newcommand{\SK}{\mathcal{K}}

\definecolor{wacvblue}{rgb}{0.21,0.49,0.74}
\usepackage[pagebackref,breaklinks,colorlinks,allcolors=wacvblue]{hyperref}

\usepackage[capitalize]{cleveref}
\crefname{section}{Sec.}{Secs.}
\crefname{section}{Section}{Sections}
\crefname{table}{Table}{Tables}
\crefname{table}{Tab.}{Tabs.}
\crefname{appendix}{Suppl.}{Suppls.}
\Crefname{appendix}{Supplement}{Supplements}

\begin{document}

\title{Cycle-Consistent Multi-Graph Matching for Self-Supervised Annotation of C.~Elegans}

\author{
	Christoph Karg\thanks{${}^{,\dag}$ Equal contribution; authors order may be altered in citations.}$^{\phantom{*},1,2}$ \quad
	Sebastian Stricker$^{*,4}$ \quad
	Lisa Hutschenreiter$^{4}$ \\
	Bogdan Savchynskyy$^{\dag,4}$ \quad
	Dagmar Kainmueller$^{\dag,1,2,3}$ \\\\
	$^1$Max-Delbrueck-Center for Molecular Medicine in the Helmholtz Association \quad
	$^2$Helmholtz Imaging \\
	$^3$University of Potsdam \quad
	$^4$Heidelberg University
}
\maketitle

\begin{abstract}
In this work we present a novel approach for unsupervised multi-graph matching, which applies to problems for which a Gaussian distribution of keypoint features can be assumed. 
We leverage cycle consistency as loss for self-supervised learning, and determine Gaussian parameters through Bayesian Optimization, yielding a highly efficient approach that scales to large datasets. 
Our fully unsupervised approach enables us to reach the accuracy of state-of-the-art supervised methodology for the biomedical use case of semantic cell annotation in 3D microscopy images of the worm C.~elegans. 
To this end, our approach yields the first unsupervised \emph{atlas} of C.~elegans, i.e. a model of the joint distribution of all of its cell nuclei, without the need for any ground truth cell annotation. 
This advancement enables highly efficient semantic annotation of cells in large microscopy datasets, overcoming a current key bottleneck. Beyond C.~elegans, our approach offers fully unsupervised construction of cell-level atlases for any model organism with a stereotyped body plan down to the level of unique semantic cell labels, and thus bears the potential to catalyze respective biomedical studies in a range of further species. 
\end{abstract}
%
%
%

%
\begin{figure*}[t]
    \centering        \includegraphics[width=0.37\linewidth]{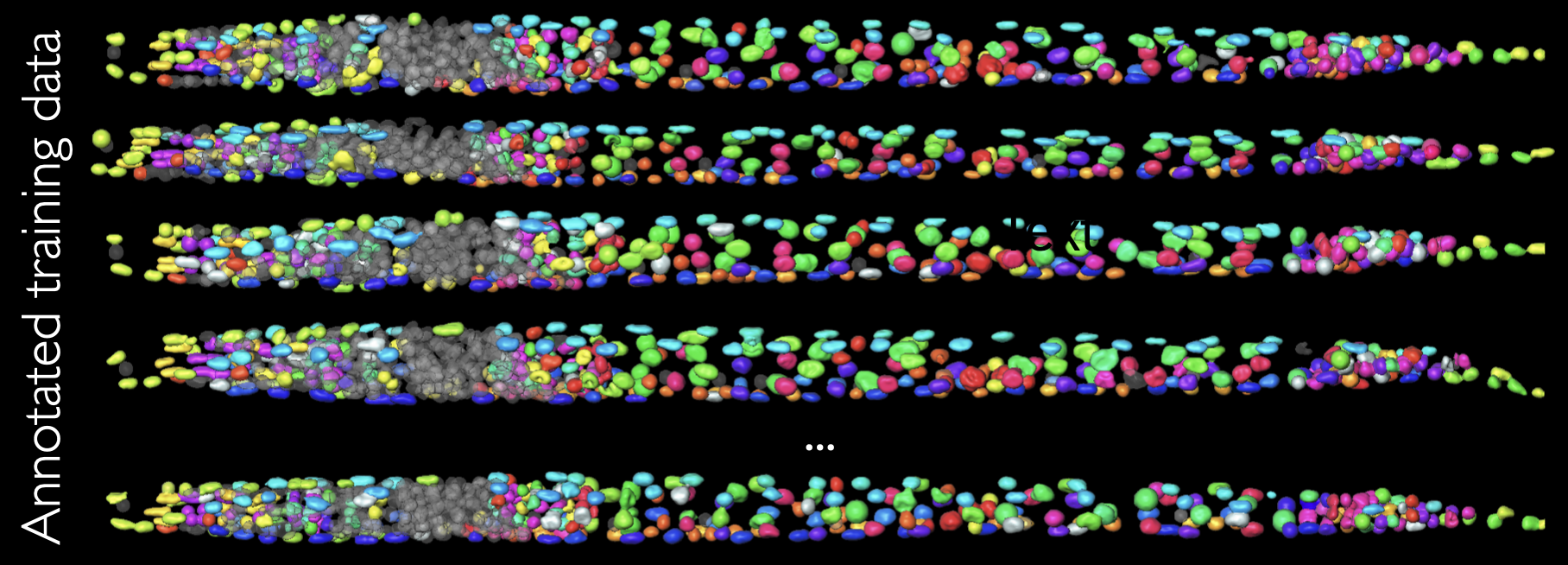}
\includegraphics[width=0.62\linewidth]{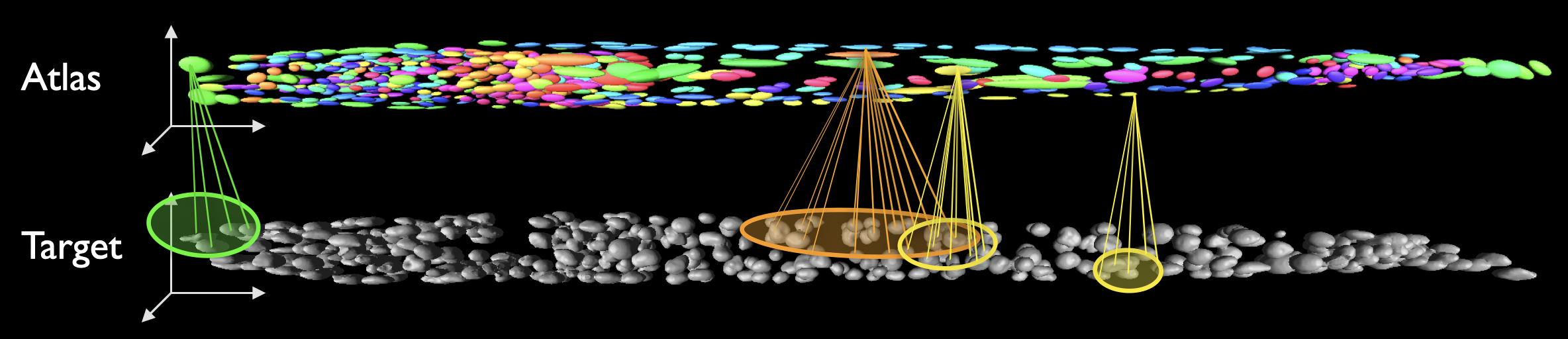}
    \caption{\textbf{Traditional approach: Supervised worm-to-atlas matching.} Left: Five exemplary worms, consistently composed from 558 cells; each cell is expert-annotated with its unique semantic name, indicated by 558 distinct colors (note, not all colors can be distinguished by eye). Such training data serves for supervised learning of a statistical atlas (top right, atlas labels illustrated by 3D ellipsoids representing the positional covariances $\Sigma_i^{\text{cen}}$ of individual cells). A target worm (bottom right, cell instance segmentation) can then be semantically annotated by solving a graph matching problem to optimally assign atlas labels to target cell instances. For computational efficiency, the problem can be sparsified by restricting the set of possible assignments for each atlas label to a reduced subset of segments (colored ovals). 
    }
    \label{fig:supervised_gaussians}
\end{figure*}
\begin{figure*}[!h]
    \centering
    \includegraphics[width=\linewidth]{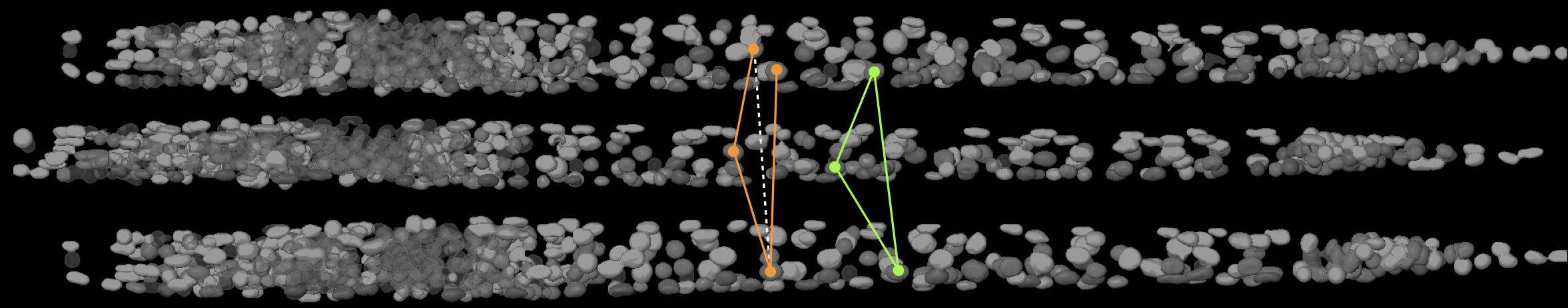}
    \caption{\textbf{Our approach: Cycle-consistent multi-graph matching for self-supervised atlas learning.} Given a set of worms (instance segmentations but no semantic labels; three examples shown), establishing \emph{cycle consistent}  correspondences for all cells across all worms yields cell cliques that effectively serve to replace semantic annotations (see green example for one cell). Determining cell cliques can be phrased as a \emph{multi-graph matching} problem, which extends pairwise matching problems to include cycle consistency constraints. Cycle consistency is necessary but not sufficient for correctness. Vice-versa, inconsistency entails error (orange example). We thus leverage cycle consistency as self-supervisory signal to learn the parameters of a graph matching objective via Bayesian Optimization, yielding the first unsupervised statistical atlas of C.~elegans. This unsupervised atlas can then be plugged into a standard worm-to-atlas matching objective.  }
    \label{fig:worms_fig}
\end{figure*}

\section{Introduction}
\label{sec:intro}
Graph matching (GM) is a critical task in computer vision, where it serves to establish correspondences between keypoints across images.  Research has primarily focused on supervised GM, while unsupervised approaches—which don't require ground truth keypoint pairs—have received far less attention. This gap is significant in large-scale real-world applications where manual labeling can be forbiddingly expensive.

One such real-world application stems from the biomedical domain, concerning the nematode worm C.~elegans, a widely used model organism in biology. C.~elegans is particularly valued for its cell-level stereotypicity: 
Its consistent body plan, composed from a fixed set of cells, allows to establish cell-level correspondences across worms (See \cref{fig:supervised_gaussians,fig:worms_fig}) and thus map cell-level observations across individuals into a common reference frame, enabling comprehensive studies of cellular processes, like e.g. gene expression \cite{Liu2009cellfate,li_full-body_2024}. 
Respective GM problems for establishing cell-level correspondences in C.~elegans have been adopted by the computer vision community, constituting some of the largest benchmark GM problems to date \cite{swoboda2017study,swoboda2019convex,kahl_2024,hutschenreiterfusionmoves-2021,haller2022comparative}. 

Traditionally, cell-level correspondences in C.~elegans are obtained via semantic annotations of individual cells with their unique biological names in 3D microscopy images. 
The most successful computational approach addressing this challenge to date 
is based on the construction of a statistical (Gaussian) atlas of cell nuclei from a given training set of worms, requiring ground truth annotations of semantic cell names \cite{long_3d_2009,golland_active_2014,li_full-body_2024} (see \cref{fig:supervised_gaussians}). 
Once such an atlas is built, new, unlabeled worms can be annotated via GM  \cite{haller2022comparative,hutschenreiterfusionmoves-2021}, allowing semantic cell names to be transferred from the atlas to the target worm. 
However, a major drawback of such supervised approaches is their reliance on ground truth semantic annotations, requiring manual expert labeling, which is an expensive, forbiddingly time-consuming, and error-prone process.

To overcome the reliance on ground truth semantic annotations, we here propose the first \emph{fully unsupervised} approach for semantic cell annotation in C.~elegans, eliminating the need for manual labeling. More specifically, we construct a statistical atlas of C.~elegans from a training set of \emph{only cell instance segmentations} (which can be obtained at viable accuracy by state-of-the-art cell \emph{instance} segmentation models  \cite{cellpose_2021,startdist_weigert2022,hirsch2020auxiliary}), 
\emph{but no semantic annotations}. 

Our approach advances upon recent work on deep unsupervised multi-graph matching (MGM) \cite{tourani_unsupervised_2024}: Their method relies on a pre-trained feature extraction network, whose parameters are optimized w.r.t.\ a self-supervised discrete cycle-consistency loss. 
While their approach is viable for standard computer vision data, its reliance on a well-pre-trained backbone hampers its generalization to intricate biomedical images, in particular to our use case of highly self-similar semantic classes. 
%
%
Thus, instead of optimizing a feature extraction backbone, we leverage Bayesian Optimization (BO) to directly optimize Gaussian parameters of matching costs, which are assumed to govern keypoint features such as their centroids and pairwise offsets. 

Our unsupervised approach achieves \textbf{96.1\%} accuracy, beating the so far best-known 93\% accuracy of a \emph{supervised} baseline~\cite{li_full-body_2024}, and  is only slightly inferior to our \emph{new} state-of-the-art \emph{supervised} baseline with 96.4\% accuracy.

\noindent \textbf{In summary, we contribute:}
\begin{itemize}
    \item A novel BO framework for learning the Gaussian parameters of an unsupervised MGM objective. In particular, we introduce a new loss function tailored to this setting and identify a class of MGM methods that are inherently biased toward optimizing it.
    \item The \emph{first} unsupervised statistical atlas of C.~elegans established with our approach from a set of 3D microscopy images;
    \item A new \emph{supervised} baseline for semantic annotation of C.~elegans nuclei that beats the previous (supervised) state-of-the-art by a large margin;
    \item A comparative evaluation of our unsupervised approach against this supervised baseline, showing that unsupervised achieves comparable accuracy and thus effectively dissolving the long-standing bottleneck of obtaining semantic ground truth annotations of C.~elegans cell nuclei.
\end{itemize}
Our code is available at \url{https://github.com/vislearn/cellmatch}.

\section{Background}
To put our contribution into context, in the following, we first briefly review the mathematical description of a cell-level atlas of C.~elegans, further referred as \emph{atlas}, and formulate the problem of \emph{worm-to-atlas} matching, both as given in~\cite{golland_active_2014}. Subsequently, we motivate and formulate the problem of \emph{worm-to-worm matching} and \emph{worm multi-matching} as performed in~\cite{swoboda2019convex}.

\subsection{Atlas-based annotation of C.~elegans} 
\label{sec:atlas-to-worm} 
\noindent\textbf{Atlas as multivariate Gaussian. } 
An atlas serves as a statistical model that captures the spatial distribution and structural variability of nuclei across multiple samples.
Following the atlas representation suggested in \cite{golland_active_2014} we model it as a multivariate Gaussian distribution, where an atlas nucleus is represented by an ellipsoid with sampled centroid and sampled main axis radii.

Let the finite set~$\mathcal{L}$ be the set of \emph{atlas nuclei} or \emph{labels}. Each nucleus $i\in\mathcal{L}$ is associated with two 3D Gaussian distributions $\SN(\bar{x}^{\text{cen}}_i,\Sigma^{\text{cen}}_i)$ for the position of its center $x^{\text{cen}}_i$ and $\SN(\bar{x}^{\text{rad}}_i,\Sigma^{\text{rad}}_i)$ for the three principal axis radii $x^{\text{rad}}_i$. Here, $\bar{x}^{\text{cen}}_i,\bar{x}^{\text{rad}}_i\in\R^3$ are the mean position and radii vectors and $\Sigma^{\text{cen}}_i$ and $\Sigma^{\text{rad}}_i$ are the respective $3\times 3$ covariance matrices.
Furthermore, for any two nuclei $i,j\in\mathcal{S}$ it is assumed that the \emph{offset vector} $x_{ij}^{\text{off}} = x_i^{\text{cen}} - x_j^{\text{cen}} \in\mathbb{R}^3$ is distributed as $\SN(\bar{x}_{ij}^{\text{off}},\Sigma^{\text{off}}_{ij})$. These distributions model correlated movement of different nuclei.
In a supervised setting, all parameters of an atlas can be straightforwardly estimated from annotated samples as empirical means and covariances~\cite{golland_active_2014}.

\noindent\textbf{C. elegans annotation by worm-to-atlas matching. }
For a new target worm, we aim to label each of its nuclei with its unique biological name. 
Given an atlas (however obtained), as proposed in~\cite{golland_active_2014}, this can be achieved by solving a GM problem~\cite{haller2022comparative} we term 
 \emph{worm-to-atlas matching}. 
%
%
This problem treats image segments (nuclei) of the target worm as samples drawn from the distribution defined by the atlas, and aims to find a correspondence between target segments and atlas labels that maximizes the likelihood under the atlas model~\cite[Sec.~2]{golland_active_2014}.

The GM formalization considers two finite sets, the set of (atlas) \emph{labels} $\SL$ and the set of \emph{segments} or \emph{nuclei} $\mathcal{S}$. W.l.o.g.\ we assume these sets to contain integer numbers, so the relation $i<j$ is defined for any $i,j$ from $\SL$ or $\mathcal{S}$.
Elements of $\SL$ and $\mathcal{S}$ must be matched to each other, as visualized in \cref{fig:supervised_gaussians}. 
For each $i\in\SL$ and $s\in\mathcal{S}$ the \emph{linear} cost $c_{is}$ of their matching is given, as well as the \emph{quadratic} cost $c_{is,jt}$ for any pair $i,j\in\SL$, $i\neq j$ if it is matched to any pair $s,t\in\mathcal{S}$, $s\neq t$. 
In our application, we calculate costs via the  \emph{Mahalanobis distance} measure \cite{mahalanobis_DeMaesschalck2000} and define linear costs as:
\begin{align}\label{equ:cost-c-is}
c_{is}:= & \lambda^\text{cen}d^{\text{cen}}(i,s) + \lambda^\text{rad}d^{\text{rad}}(i,s)\,, \ \textnormal{where} \\
d^{\text{cen}}(i,s)= & (\bar{x}_i^{\text{cen}} - x_s^{\text{cen}})^T (\Sigma^{\text{cen}}_i)^{-1} (\bar{x}_i^{\text{cen}} - x_s^{\text{cen}})\,,\\
d^{\text{rad}}(i,s)= & (\bar{x}_i^{\text{rad}} - x_s^{\text{rad}})^T (\Sigma^{\text{rad}}_i)^{-1} (\bar{x}_i^{\text{rad}} - x_s^{\text{rad}})\,.
\end{align}
The quadratic costs are defined as
\begin{multline} \label{equ:cost-c-is-c-st}
c_{is,jt} := \lambda^{\text{off}}d^{\text{off}}(i,j,s,t)\\ 
:= \lambda^{\text{off}} (\bar{x}_{ij}^{\text{off}} - x_{st}^{\text{off}})^T (\Sigma^{\text{off}}_{ij})^{-1} (\bar{x}_{ij}^{\text{off}} - x_{st}^{\text{off}})\,,
\end{multline}
where $\lambda^{\text{cen}}, \lambda^{\text{rad}}, \lambda^{\text{off}} >0$ are weight parameters controlling the relevance of the respective features.
The goal of GM is to find a \emph{matching} or \emph{assignment} between elements of sets $\SL$ and $\mathcal{S}$ that minimizes the total cost~\eqref{equ:gm}. This matching must satisfy the \emph{uniqueness constraint}~\eqref{equ:gm_constr}, that is, each element in $\SL$ must be assigned to \emph{at most one} element in $\mathcal{S}$ and vice versa. \emph{At most one} implies that some elements may remain unassigned, which in our case may happen due to missing or superfluous segments. In total, GM is formulated as the following integer program:
%
\begin{align}\label{equ:gm}
    \min_{x\in \{0,1\}^{|\SL|\times|\mathcal{S}|}}
        \sum_{i\in\SL\atop s\in\mathcal{S}}C_{is}x_{is}+ 
        \sum_{i,j \in \mathcal{L}\atop i<j}\sum_{s,t \in \mathcal{S}\atop s<t} 
        c_{is,jt}\cdot x_{is}x_{jt} \\
   \mathrm{s.t.}
   \begin{cases}\label{equ:gm_constr}
        \forall i \in \mathcal{L}: \sum_{s \in \mathcal{S}} x_{is} \leq 1,\\
        \forall s \in \mathcal{S}: \sum_{i \in \mathcal{L}} x_{is} \leq 1\,.
    \end{cases}
\end{align}
Here the $x_{is}$ are binary variables that indicate whether atlas label $i$ is assigned to target image segment $s$, and $C_{is} = c_{is} - c_0$. Here, $c_0>0$ is a \emph{unassignment cost} constant, subtracted to allow for negative assignment costs, thus avoiding a trivial \emph{nothing-assigned} solution.

Should the quadratic costs $c_{is,jt}$, be equal to zero for all keypoint four-tuples, the GM problem~\eqref{equ:gm}-\eqref{equ:gm_constr} reduces (\cite{haller2022comparative}) to the polynomially solvable \emph{linear assignment} problem~\cite{burkard2012assignment}.

One may \emph{sparsify} the GM problem~\eqref{equ:gm}-\eqref{equ:gm_constr} by forbidding any a-priori-known implausible matchings of segment $s$ to label $i$ by assigning the respective cost $c_{is}$ an infinite value.
This allows GM solvers that can leverage sparsity to deal with larger sets $\SL$ and $\mathcal{S}$. 
The GM problem is NP-hard, but for sparse problem instances with $|\SL|,|\mathcal{S}|\approx 500-1000$ as in our case, there are methods able to find fairly accurate solutions in seconds, see the benchmark~\cite{haller2022comparative}. In our work, we employ the winner~\cite{hutschenreiterfusionmoves-2021} of this benchmark and we apply the same sparsification for worm-to-atlas matching as described in~\cite{golland_active_2014}.

Contrary to the general case GM problem, the linear assignment problem is efficiently solvable for $\SL,\mathcal{S}\approx 1000$ even for \emph{dense} costs $c_{is}$~\cite{crouse2016}. 

\subsection{C. elegans annotation by multi-graph matching}
\label{sec:mgm}
As a first step towards unsupervised annotation, we consider the case where label-specific nucleus resp. nucleus-pair covariance matrices $\Sigma_i^{\{\text{cen}, \text{rad}\}}$ resp. $\Sigma_{ij}^\text{off}$ are unavailable.
Instead, we assume that \emph{cross-nuclei} (i.e. label-independent) covariance matrices $\Sigma^{\{\text{cen}, \text{rad}, \text{off}\}}$ are fixed across all nuclei resp. nuclei pairs.
For instance, they can be assigned to identity matrices.

With this information, correspondences between target worms can still be determined~\cite{swoboda2019convex}: For a pair of worms, we can establish a \emph{worm-to-worm matching} problem, taking the same general form as the worm-to-atlas matching problem described in Sec.\ \ref{sec:atlas-to-worm}, by replacing any label-specific covariance matrix with its cross-label counterpart. 
Also, each label-specific mean centroid $\bar{x}^{\text{cen}}_i$ of the atlas is replaced by the actual centroid $x^{\text{cen}}_i$ of the worm to be matched to the target worm, and we proceed analogously for radii and offset vectors.
The worm-to-worm matching will serve as our main example of a \emph{pairwise GM subproblem} in what follows.

Generalizing beyond pairs of worms, the problem considered in~\cite{swoboda2019convex} is to achieve consistent pairwise matchings for a \emph{set of worms}. 
This constitutes a \emph{worm multi-matching problem}, an instance of MGM, which naturally extends the GM problem~\eqref{equ:gm}-\eqref{equ:gm_constr} to $N > 2$ keypoint (in our case \emph{segmented nuclei}) sets $(\mathcal{S}^1, \dots, \mathcal{S}^N)$. 
%
%
The MGM objective is to find a \emph{cycle consistent} correspondence between the keypoint sets, that minimizes the total cost over all pairwise GM subproblems corresponding to the $N \choose 2$ pairs of keypoint sets.
Cycle consistency, as visualized in \cref{fig:worms_fig}, means transitivity of matchings, i.e., for any sequence of keypoints $s_1, ... ,s_k$ over $k$ different keypoint sets, $s_i$ being matched to   $s_{i+1}$ for all $i<k$ implies that $s_k$ is matched to $s_1$. Cycle consistency across all triplets of keypoint sets is sufficient to ensure  cycle consistency across all keypoint sets~\cite{swoboda2019convex}.   
Cycle consistency of triplets can be captured as a set of additional linear  constraints (see~\cite{swoboda2019convex} for details), which link the otherwise independent pairwise GM subproblems into an MGM problem.

\subsection{Used MGM solvers}\label{sec:used-mgm-solvers}
Sparse MGM problems, i.e., MGMs with sparse GM subproblems as in our case, are solved in a highly efficient manner by the recently published state-of-the-art algorithm \cite{kahl_2024}.
This algorithm has two modes:
\begin{itemize}
    \item \emph{Direct} mode, where minimization of the original MGM objective is addressed, subject to all constraints. The latter consists of uniqueness constraints~\eqref{equ:gm_constr} as well as constraints ensuring cycle consistency. 
    In this mode, the MGM algorithm~\cite{kahl_2024} internally leverages the Fusion Moves algorithm~\cite{hutschenreiterfusionmoves-2021} to solve GM subproblems.
    \item \emph{Synchronization} mode, where it first solves all pairwise GM problems \emph{independently} (\eg, with~\cite{hutschenreiterfusionmoves-2021}), causing a cycle-\emph{inconsistent} matching. Afterwards, it finds its cycle-consistent approximation by changing a minimal number of matchings~\cite[Sec.~7]{kahl_2024}.
    This second step itself can be seen as MGM, where each pairwise GM subproblem has only linear costs 
   \begin{equation}\label{equ:synchronization-cost}
      C_{is}
      =
      \left\{ 
        \begin{array}{ rl }
          -1,&  i\ \text{is matched to}\ s\,, \\
          \infty, & c_{is}=\infty\ \text{and sparsification is ON}\\
          0, &\text{ otherwise }\,. \\
        \end{array}
      \right. 
    \end{equation}
    and is, therefore, equivalent to the efficiently solvable linear assignment problem. Due to this, the direct solver applied to this MGM runs several orders of magnitude faster than in the general case.

    Another advantage of the synchronization mode is the possibility to drop the cost sparsification at the second step of the algorithm. This can be done by ignoring the second line in~\eqref{equ:synchronization-cost}. As a result, the final solution may
    contain initially forbidden matchings. Given that the sparsification in our case serves mainly to reduce the computational complexity caused by quadratic costs, dropping it enables a slight improvement of the final matching accuracy; see \cref{fig:compare_modes} in \cref{sec:ablations-and-results}.
    Consequently, we refer to the two synchronization modes considered as \emph{sparse} and \emph{dense}. 
\end{itemize}
It is worth mentioning that there exist a number of direct and synchronization-based MGM solvers in addition to~\cite{kahl_2024}. However, they seem to be inferior to the latter in terms of accuracy and speed, according to the evaluation in~\cite{kahl_2024}.

\section{Unsupervised Atlas Learning via Bayesian Optimization}
\label{sec:unsupervised-atlas-learning}


As shown in \cite{tron2017fast}, a cycle consistent solution of an MGM problem naturally yields \emph{cliques} of matched keypoints, comprising (at most) one nucleus per worm.

Recall that each nucleus of the supervised atlas is modeled via 3D Gaussian distributions (see \cref{sec:atlas-to-worm}) over a set of sample nuclei with the same biological ground truth label.
Given an unsupervised MGM solution, we can use its cliques instead of these annotated sample sets to build an \emph{unsupervised atlas}. 
In \cref{sec:unsupervised-atlas-labels}, we describe how to assign ground truth labels to the cliques to allow evaluation and comparison with the supervised case.

The quality of this unsupervised approach depends critically on the parameters that define the cost function of the underlying MGM problem. We therefore examine learning of these parameters in the remainder of this section.

\noindent\textbf{Parameters of the worm multi-matching problem to be learned}
are the cross-nuclei covariance matrices $\Sigma^{\{\text{cen}, \text{rad}, \text{off{}}\}}$ and the cost weights $\lambda^\text{\{cen, rad, off\}}$, see 
\crefrange{equ:cost-c-is}{equ:cost-c-is-c-st}. 
We further sparsify our \emph{worm-to-worm matching} problems using three hyper-parameters $K_\text{min}, \tau^\text{cen}, \tau^\text{rad}$: We forbid all assignments where the centroid distance $d^{\text{cen}}(i,s)$ is above a certain threshold $\tau^{\text{cen}}>0$, and proceed analogously with a radii threshold $\tau^{\text{rad}}>0$. While precomputing all possible linear costs $c_{is}$, we allow at least the $K_\text{min}$ lowest cost assignments of every nuclei. 
A further parameter is the unassignment cost $c_0$ which incentivizes non-trivial matchings. 
It effectively controls how many nuclei remain \emph{unmatched}. We match as many nuclei as possible by setting the unassignment cost $c_0$ to a large constant, exceeding any finite $c_{is}+c_{jt}+c_{is,jt} $.


\subsection{Loss function.}\label{sec:loss-function}
Cycle consistency as described in \cref{sec:mgm} has been shown to serve as a self-supervisory loss in related works on deep GM~\cite{tourani_unsupervised_2024}. 

In a nutshell, one selects a batch of $N_\text{learn}$ keypoint sets (\emph{worms} in our case) to be matched, solves the respective $N_\text{learn} \choose 2$ pairwise GM subproblems
independently from each other and counts the number of inconsistent keypoint matching triples, see~\cite{tourani_unsupervised_2024} for a rigorous definition. This number is known as the \emph{discrete cycle loss} that has to be minimized during learning. 

Although this approach aims to decrease the number of inconsistencies, there is no cycle consistency guarantee 
when all pairwise GM subproblems are solved independently, 
so an MGM algorithm must be used on top to obtain a consistent multi-matching. 
When testing the different MGM solver modes to obtain this multi-matching, we have found that the synchronization modes attain a higher matching accuracy than the direct one, see \cref{fig:compare_modes} in \cref{sec:ablations-and-results}, although the latter shows better results in terms of the total cost value. 
Our hypothesis to this behavior is that the discrete cycle loss correlates with the objective of the synchronization modes
and, as a result, the learned parameters are biased towards synchronization algorithms. 

To check this hypothesis empirically, we substituted the discrete cycle loss with the \emph{synchronization} objective~\eqref{equ:synchronization-cost}, further referred to as \emph{synchronization loss}. 
As mentioned in \cref{sec:used-mgm-solvers}, an approximate computation of this objective with the MGM algorithm~\cite{kahl_2024} is very efficient, see \cref{table:runtimes}.
In this way our learned parameters are \emph{explicitly} biased towards synchronization MGM algorithms, which resulted in a slight improvement of the matching accuracy, compared to the usage of the discrete cycle loss, see \cref{table:ablation_loss} in \cref{sec:ablations-and-results}. 
To the best of our knowledge, this is \emph{the first usage} of the synchronization loss for self-supervised learning of (M)GM algorithms.

\subsection{Parameter learning via Bayesian optimization}\label{sec:BO-learning}
\noindent\textbf{Classification of learned parameters.} Parameters of the MGM problem as described above can be split into three groups:
\begin{enumerate}
    \item \emph{Linear cost} parameters, which include the covariance matrices $\Sigma^\text{cen}$ and  $\Sigma^\text{rad}$. Note, when treating them as learnable parameters, their  weighting coefficients $\lambda^\text{cen}$ and $\lambda^\text{rad}$ become redundant and can be discarded. To further simplify the problem, we consider only diagonal covariance matrices, since variance mainly differs in the three body axes, the latter being pre-determined by unsupervised rigid worm alignment. Two diagonal $3\times 3$ matrices give in total 6 parameters.
    \item \emph{Sparsity} parameters $K_\text{min}, \tau^\text{cen}, \tau^\text{rad}$, three in total.
    \item \emph{Quadratic cost} parameters, defined by the covariance matrix $\Sigma^\text{off}$. For the same reasons as above, we assume this matrix to be diagonal and discard the weighting coefficient $\lambda^\text{off}$. As above, the matrix is defined by its three diagonal elements.
\end{enumerate}
All in all, we have 12 parameters to learn.

\noindent\textbf{Optimization method.} 
Whereas the linear and quadratic cost parameters could potentially be learned with a gradient-based technique, this is absolutely inadequate for the sparsity parameters.
Additionally, the non-convexity of the optimization problem in the low-dimensional parameter space makes the gradient-based techniques prone to get stuck in local minima.
\footnote{Note that \cite{tourani_unsupervised_2024} does not face these issues in a similar context, as it considers learning of \emph{small-sized dense} GM problems whose costs are determined by a deep neural network.}


To address these challenges, we adopt a gradient-free BO approach~\cite{snoek2012practicalbo}, which models the objective as a random function with a prior distribution. A posterior is iteratively estimated by evaluating the objective over a sufficiently large search space, and the resulting posterior defines an acquisition function that guides the selection of the next evaluation point. We implement BO using the Optuna framework~\cite{optuna_2019}, employing its default as well as multi-objective Tree-structured Parzen Estimator, see~\cite{bergstra2011_TPE} and~\cite{Ozaki_Multi_TPE_2020} respectively.

To the best of our knowledge, this is the first application of BO for self-supervised learning of (M)GM algorithms.

\noindent\textbf{Optimization pipeline.}
To enable BO at feasible computational cost, we select a subset of $N_\text{learn}=15$  worms. For these worms, we solve all ${N_\text{learn} \choose 2}$ worm-to-worm pairwise GM problems and compute the loss from the resulting matchings.

Learning all 12 parameters jointly with BO would be prohibitively slow. We therefore divide the learning procedure into three stages:
\begin{enumerate}
    \item \emph{Linear costs.}  
    We first optimize the \emph{six} linear cost parameters, discarding quadratic terms and treating the matching as a dense linear assignment problem. Removing quadratic costs drastically reduces the problem size and yields an efficiently solvable problem.
    
    \item \emph{Sparsity parameters.}  
    Next, we optimize the \emph{three} sparsity parameters, still without quadratic costs. The goal is to obtain GM instances that are as sparse as possible so that later optimization with quadratic terms remains tractable. We employ the multi-objective Tree-structured Parzen Estimator~\cite{Ozaki_Multi_TPE_2020} to jointly minimize the overall loss and the average number $\bar{n}_\text{lin}$ of allowed assignments across all ${N_\text{learn} \choose 2}$ pairs. As these objectives conflict, BO yields a Pareto set~\cite{pareto}. From this set, we retain only solutions with $\bar{n}_\text{lin} < 12000$, further restrict to those within $0.05\%$ of the lowest loss among them, and finally choose the candidate with minimal $\bar{n}_\text{lin}$.
    
    \item \emph{Quadratic costs.}  
    Finally, we reintroduce quadratic costs and optimize the remaining \emph{three} covariance parameters of $\Sigma^\text{off}$.
\end{enumerate}

\subsection{Construction of the unsupervised atlas}\label{sec:unsupervised-atlas-labels}
Although the unsupervised atlas construction is in a nutshell the same as the supervised one, there are two differences, which we describe below.

First, the parameters $\lambda^\text{\{cen,rad,off\}}$ are computed from the respective covariance matrices $\Sigma^\text{\{cen,rad,off\}}$ obtained during learning (see \cref{sec:unsupervised-atlas-learning}). This is done by
setting $\lambda^\text{\{cen,rad,off\}}:=\Vert\text{diag}(\Sigma^\text{\{cen,rad,off\}})\Vert$, where $\text{diag}(\Sigma)$ stands for the vector of diagonal elements of the matrix $\Sigma$. To maintain compatibility with the fixed unassignment parameter $c_0$ we further norm the vector $(\lambda^\text{cen},\lambda^\text{rad},\lambda^\text{off})$ such that its length is equal to the length of the vector $(1,1,1)$.

The second difference is related to the alignment defined in \cref{sec:alignment} below. Here, we have found it unnecessary to perform any additional steps beyond those already performed to compute a consistent MGM solution.

\noindent\textbf{Assigning ground truth labels to an unsupervised atlas.}
To evaluate the accuracy of the unsupervised atlas, we have to assign ground truth labels to the cliques obtained by the MGM. Let $\SK$ be the set of cliques and $\SL$ the set of ground truth labels. First we select the clique $k\in\SK$ that contains the overall maximum occurrence of some ground truth label $l \in \SL$. We assign the induced label $l$ to \emph{all} nuclei in $k$. The clique $k$ is then removed, as well as all nuclei having the ground truth label $l$ in all other cliques. The process is repeated until all labels in $\SL$ have been assigned. 
This process may result in unlabeled cliques. These are nevertheless included into the atlas and all matchings to them later on are treated as incorrect. 

\subsection{Additional processing} \label{sec:alignment}

\noindent\textbf{Pre-processing. }As an input to our method, we require three-dimensional images of C.~elegans with instance segmentation on their nuclei.
As in~\cite{golland_active_2014}, the segmented nuclei are approximated with ellipses using principal component analysis and the body axes and barycenters of all worms are rigidly aligned. 
As linear costs $c_{is}$ directly depend on the actual centroid coordinates $ x_i^{\text{cen}} $, $ x_s^{\text{cen}} $, see \cref{equ:cost-c-is}, having a good alignment is highly beneficial.

%


\noindent\textbf{Re-alignments.}
To further improve the quality of both worm-to-worm and worm-to-atlas matching, we apply additional \emph{re-alignment} steps following the approach of~\cite{golland_active_2014}.

Given a fixed matching between a worm and an atlas, an optimal affine transformation of the worm can be estimated via least squares, minimizing the total matching cost. In our setting, we apply this procedure by selecting a single worm as a target (treated as a pseudo-atlas) and re-aligning all training and test worms to it using worm-to-worm matching.

To identify a suitable target worm, we first solve all pairwise GM problems on the training set using dense and unlearned linear costs, i.e., setting $\Sigma^\text{\{cen,rad\}} = I$ and ${\lambda^\text{\{cen,rad\}} = 1}$. The resulting inconsistent multi-matching is then approximated with a consistent one using the \emph{dense synchronization} mode of the employed MGM solver, see \cref{sec:used-mgm-solvers}. The worm that retains the largest number of consistent matchings across the synchronization result is selected as the target for re-alignment.

Re-alignment is performed twice: once before and once after parameter learning via BO.
For the initial re-alignment, we use the dense unlearned linear costs as described above. 
The second re-alignment, in contrast, leverages the fully learned model, including quadratic and sparse costs, and is applied to both training and test worms. 
In both cases, we perform seven re-alignment iterations to ensure convergence.

\noindent\textbf{The scalability of our approach} is determined by two factors: the number $N_{\text{learn}}$ of worms used for parameter learning and the number of cells in each worm. 
The first limitation can be mitigated through parallel, \emph{independent} computation of all pairwise GM subproblems, which constitute the main bottleneck of the training procedure. The second factor, the number of cells, highlights the importance of sparsifying the GM subproblems:
The used GM solver~\cite{hutschenreiterfusionmoves-2021} has iteration complexity $O(|\mathcal{S}| \bar{L}^2)$, with $\mathcal{|S|}$ being the number of cells per worm, and $\bar{L}$ -- the average number of allowed assignments per cell. 
As such, if $\bar{L}$ is kept bounded as $\mathcal{|S|}$ grows, the iteration runtime scales only linearly with $\mathcal{|S|}$.

\section{Empirical evaluation}\label{sec:ablations-and-results}

\noindent\textbf{Dataset. }
\label{sec:dataset}
We showcase our approach on the dataset~\cite{li_2023_dataset} containing 3D light microscopy images of C. elegans at the L1 larval stage. Each worm contains 558 nuclei with ground truth instance segmentations as well as full semantic annotations, see~\cite{li_full-body_2024}. 
The dataset consists of one training set of 100 worms and two test sets of 100 worms each. To ensure a direct comparison to~\cite{li_full-body_2024}, we use the second test set in our experiments.



\noindent\textbf{(Supervised) (pre-)atlas accuracy.}
We evaluate the worm-to-atlas matching using \emph{atlas accuracy}, defined as the fraction of correctly classified nuclei.
We explicitly add the word \emph{supervised} to distinguish the respective accuracy in the supervised case.
For the unsupervised atlas, we assign ground truth labels to it as detailed in \cref{sec:unsupervised-atlas-labels}.


We also use the process of assigning ground truth labels described in \cref{sec:unsupervised-atlas-labels} to evaluate accuracy of the MGM solution \emph{prior} to building an atlas. In this case, the nuclei of unlabeled cliques are treated as \emph{unmatched}, the other nuclei are assigned their clique labels. We refer to this as \emph{pre-atlas accuracy} below.

\subsection{Ablation study}
Since our approach is self-supervised, we performed all ablation  on the training dataset. 

\noindent\textbf{Model complexity (\cref{table:ablation_model}).}
To confirm that learning the cross-nuclei covariances and including quadratic cost into our GM problem is worthwhile, we compare the accuracy achieved by our full pipeline against both \emph{learned} linear and \emph{unlearned} (\ie, $\Sigma^\text{\{cen,rad,off\}} = I$ and $\lambda^\text{\{cen,rad,off\}} = 1$) quadratic models.

\begin{table}
   \centering
\caption{Ablation study on pipeline simplifications. For \emph{linear dense}, we only ran step 1 of our learning pipeline (\cref{sec:BO-learning}), allowing all assignments. \emph{Linear sparse} was furthermore sparsified via step 2. For both approaches, the worm-to-atlas matching problems used to calculate atlas accuracy were also modeled as sparse and dense problems respectively.
\emph{Unlearned quadratic} uses unit covariances ($\Sigma^\text{\{cen,rad,off\}} = I$ and $\lambda^\text{\{cen,rad,off\}} = 1$) and is as sparse as \emph{linear sparse}.
}
   \begin{tabular}{@{}lcc@{}}
     \toprule
     Ablation & Pre-Atlas Acc. & Atlas Acc. \\
     \midrule
      Linear dense & $0.949 \pm 0.033 $ & $ 0.955 \pm 0.031$\\
      Linear sparse & $0.949 \pm 0.033 $ & $ 0.956 \pm 0.030$\\
      Unlearned quadratic & $0.944 \pm 0.027 $ & $ 0.964 \pm 0.018$\\
      \midrule
      Full pipeline & $0.966 \pm 0.016$ & $ 0.972 \pm 0.015$\\
     \bottomrule
   \end{tabular}
   \label{table:ablation_model}
\end{table}

\noindent\textbf{Impact of loss function (\cref{table:ablation_loss}).} Here we show a slight improvement of our \emph{synchronization loss} over the \emph{discrete cycle loss}, see \cref{sec:loss-function}. 
The former was computed by running the MGM solver~\cite{kahl_2024} in the \emph{sparse synchronization mode}, see \cref{sec:used-mgm-solvers}. 
The respective runtime is negligible compared to the cost of solving all pairwise GM problems, see \cref{table:runtimes}.
\begin{table}
   \centering
\caption{Accuracies for the two loss functions considered.}
   \begin{tabular}{@{}lcc@{}}
     \toprule
     Loss & Pre-Atlas Acc. & Atlas Acc. \\
     \midrule
      Discrete cycle loss & $0.965 \pm 0.017 $ & $ 0.970 \pm 0.015$\\
      Synchronization loss  & $0.966 \pm 0.016 $ & $ 0.972 \pm 0.015$\\
     \bottomrule
   \end{tabular}
   \label{table:ablation_loss}
 \end{table}

\noindent\textbf{Impact of MGM solver mode (\cref{fig:compare_modes}).}
As discussed in \cref{sec:used-mgm-solvers}, the MGM solver~\cite{kahl_2024} we use, can be run in \emph{direct}
or \emph{sparse/dense synchronization} mode. As \cref{fig:compare_modes} suggests and as expected given the loss function, \emph{direct} performs worse than both synchronization modes. Furthermore, \emph{dense} synchronization has a slight edge over the \emph{sparse} one. Therefore, for all remaining results in this work, the MGM solution was generated using the \emph{dense synchronization} mode.

\begin{figure}
    \centering
    \scriptsize
    \input{figures/plot_solver_mode.pgf}
    \caption{
    Atlas accuracy as a function of the training set size for different modes of the MGM solver, see \cref{sec:used-mgm-solvers}.
    \label{fig:compare_modes}}
\end{figure}

\begin{figure}
    \centering
    \scriptsize
    \resizebox{\linewidth}{!}{\includegraphics{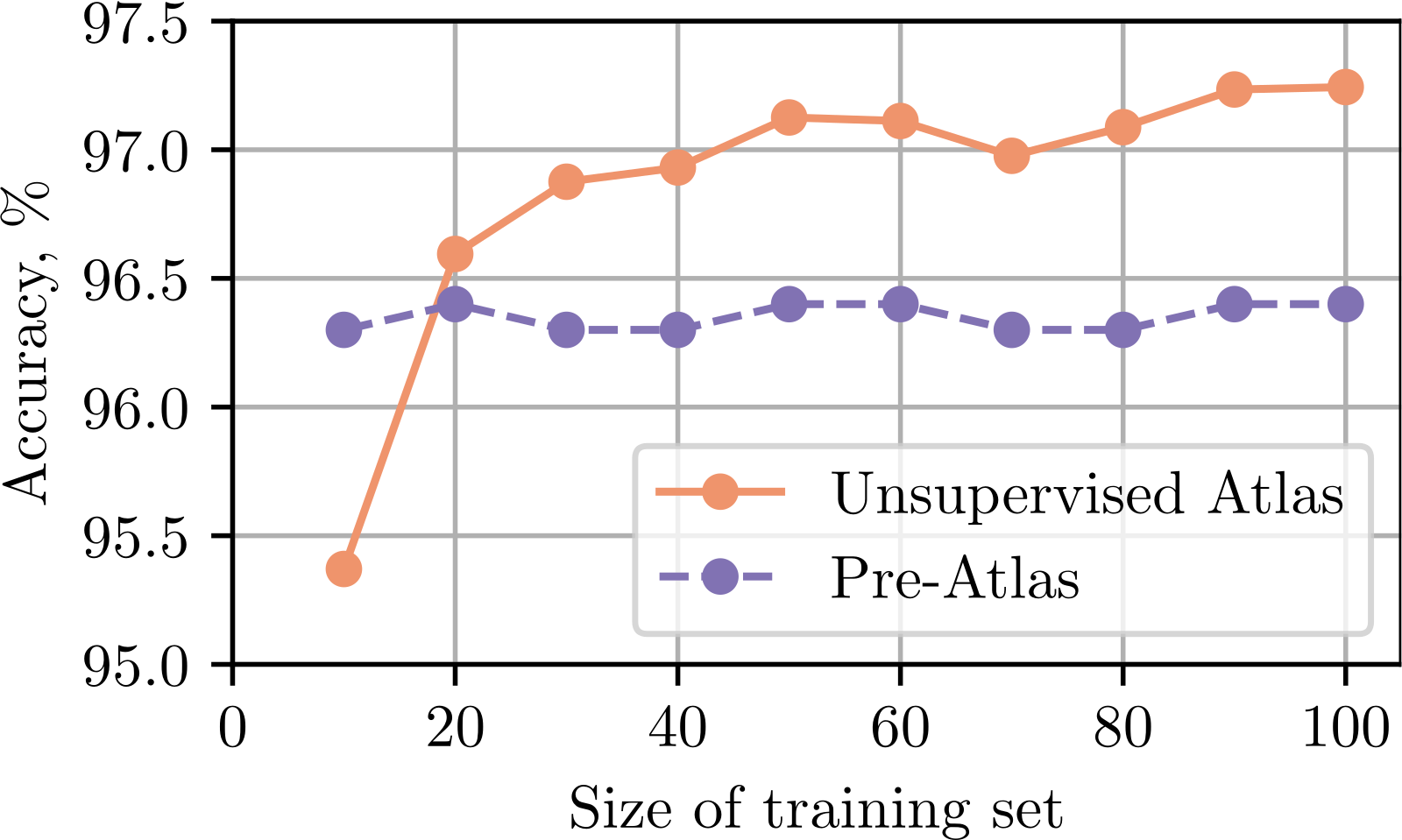}}
    \caption{
    Impact of atlas construction for varying training set sizes $N \in \{{10, 20, \dots, 100}\}$. \emph{Pre-atlas accuracy} evaluates the accuracy of the MGM solution when matching the $N$ training worms to themselves. \emph{Unsupervised atlas accuracy} shows the average accuracy when matching the same set of $N$ worms to the atlas built from their MGM solution.
    \label{fig:compare_pre_atlas_vs_atlas}}
\end{figure}

\noindent\textbf{Impact of atlas construction (\cref{fig:compare_pre_atlas_vs_atlas}).}
Since the MGM solution yields cycle-consistent matchings that cover all nuclei in the training worms, one might ask whether this output already constitutes a suitable pairwise matching -- raising the question of whether explicit atlas construction is necessary to obtain high-quality matchings.

To assess the impact of atlas construction, we evaluate, for a selection of $N \in \{{10, 20, \dots, 100}\}$ training worms (i) the \emph{pre-atlas accuracy}, \ie, the accuracy of the MGM solution on the $N$ training worms, and (ii) the atlas accuracy, \ie, the accuracy of the unsupervised atlas built from the same $N$ worms and matched back to them.

As both accuracy measures are computed on the same subset of worms, they can be directly compared, see \cref{fig:compare_pre_atlas_vs_atlas}. The results demonstrate that atlas construction significantly improves matching accuracy for $N \geq 20$. Moreover, while worm-to-atlas accuracy keeps improving with larger training sets, the pre-atlas accuracy remains nearly constant. This suggests that the benefits of additional training data are primarily realized through atlas construction.

We attribute the outperformance of the atlas to the greater modeling capacity of the cost functions used in worm-to-atlas matching, which incorporate \emph{nucleus-specific} covariance matrices. In contrast, worm-to-worm matching uses shared (nucleus-independent) covariance matrices, limiting expressiveness.

At the same time, outperformance of the pre-atlas estimate for small $N$ can be explained by the  suboptimality of statistical decisions based on maximum likelihood estimates (such as empirical means and covariances used to build the atlas) for training sets of limited size~\cite{schlesinger2022minimax}. 

\noindent\textbf{Test with automatic segmentation.}
This experiment evaluates the robustness of our method to segmentation errors. We applied StarDist~\cite{startdist_weigert2022} \emph{without} tuning it to our data. The resulting supervised accuracy is 91.6\%, and the unsupervised accuracy is 82.9\%. We found the \emph{pre-processing} step as described in \cref{sec:alignment} to be the main factor limiting performance. Detailed results and discussion are provided in \cref{suppl:stardist}.
\subsection{Comparison to supervised method}
\label{sec:results_comparison}
For the experiments in this section, we build the atlas as described in \cref{sec:atlas-to-worm} from 100 training worms, and perform atlas-to-worm matching on the 100 test worms.

\noindent\textbf{New supervised baseline (\cref{table:supervised_comparison}). }
As shown in \cref{table:supervised_comparison}, our supervised atlas construction pipeline shows better results than those of the main competing works~\cite{li_full-body_2024} and~\cite{golland_active_2014}. Although primarily based on~\cite{golland_active_2014} it differs from it in two aspects: (i) Parameters $\lambda^\text{cen} = 0.48$, $\lambda^\text{rad} = 0.34$ and $\lambda^\text{off} = 0.81$ differ from those of~\cite{golland_active_2014}. We tuned their values with BO. We assume that the overall improvement over \cite{golland_active_2014} mainly results from this change.
(ii) We implemented re-alignment as described in~\cite{li_full-body_2024}, which we found to be less expensive while still yielding good results.

In comparison to~\cite{li_full-body_2024} our pipeline is improved primarily by employing quadratic costs for GM whose parameters have been tuned with BO.
We refer to \cref{sec:supervised_atlas_construction} for finer details.

\noindent\textbf{Unsupervised vs. supervised atlas (\cref{table:supervised_comparison}, \cref{fig:compare_unsupervised}).} The accuracy of our unsupervised atlas closely matches its supervised counterpart and significantly outperforms those of~\cite{golland_active_2014} and~\cite{li_full-body_2024}. 
As~\cite{golland_active_2014} does not report accuracy on our dataset, we provide results based on our own reimplementation. In this version, we retain the original method's structure but replace its GM solver with the one from~\cite{hutschenreiterfusionmoves-2021}, which we consistently use across all experiments for a fair comparison.

We further illustrate in \cref{fig:compare_unsupervised}, how increasing the number of worms used to construct the atlas influences the final accuracy. 
In addition, \cref{fig:supervised_unsupervised_density} in the Supplement depicts the specific accuracy value of every test worm for both our supervised and unsupervised atlas.


%
\begin{figure}
    \centering
    \scriptsize
    \input{figures/plot_supervised_unsupervised_train_size.pgf}
    \caption{
    Accuracies of the unsupervised atlas as a function of the training set size, evaluated on all 100 test worms. As  an upper bound, accuracy of our supervised atlas built from 100 training worms is shown.
    \label{fig:compare_unsupervised}}
\end{figure}
\begin{table}[t]
    \centering
    \caption{Supervised- and unsupervised atlas-based C.~elegans annotation on our 100/100 train/test split. Accuracy is given as the mean over all test set worms with standard deviation.  }
   \begin{tabular}{@{}lcc@{}}
     \toprule
     Method & Accuracy\\
     \midrule
     Supervised Atlas~\cite{li_full-body_2024} &   $0.93 \pm 0.11$\\
     \midrule
     Our supervised Atlas & $0.964\pm0.019$\\
     Our unsupervised Atlas & $0.961\pm0.020$\\
     \bottomrule
   \end{tabular}
   \label{table:supervised_comparison}
 \end{table}
\noindent\textbf{Runtime analysis} is given in \cref{table:runtimes} and its caption. All computations have been performed on the same machine in a single thread mode, see \cref{suppl:runtimes} for details.

\begin{table}
   \centering
\caption{Average runtime per BO learning iteration for $N_\text{learn}$ worms. Differences in runtime for different loss functions is explained mainly by inefficiency of implementation (See \cref{suppl:runtimes}). When implemented efficiently we expect the times for both synchronization losses to be comparable to that of the discrete cycle loss and, in total, negligible compared to other computations. Similarly, we expect the runtime for Steps 1 and 2 to become much shorter than that of the Step 3. Currently, the computational overhead of (re-)building the pairwise GM problem instances after each cost update outweighs the runtime of the GM solver.}
   \begin{tabular}{@{}lcc@{}}
     \toprule
     Optimization substep & runtime (mm:ss) \\
     \midrule
      Step 1 (dense linear)       & 02:26 \\
      Step 2 (sparse linear)       & 02:18 \\
      Step 3 (sparse quadratic)       & 04:32 \\
      Discrete cycle loss     & 00:10 \\
      Synchronization loss (dense) & 00:50 \\
      Synchronization loss (sparse)& 00:01 \\
     \bottomrule
   \end{tabular}
   \label{table:runtimes}
\end{table}

\section{Conclusions}
\label{sec:discussion}
We presented a novel, scalable approach for unsupervised multi-matching based on Gaussian-distributed features. Leveraging recent advances in GM and MGM optimization -- enabled by fast and accurate solvers~\cite{hutschenreiterfusionmoves-2021,kahl_2024} -- our method scales efficiently to hundreds of keypoints across dozens or even hundreds of keypoint sets. In the task of atlas construction for C.~elegans, our approach surpasses all baselines, including supervised methods, by a substantial margin.

Future work includes integrating deep neural networks into our pipeline to replace hand-crafted linear cost features and applying the framework to other stereotypical organisms.

\FloatBarrier
\clearpage

\subsubsection*{Acknowledgments} This work was supported by the German Research Foundation projects 498181230 and 539435352.
Authors further acknowledge facilities for high throughput calculations bwHPC of the state of Baden-Württemberg (DFG grant INST 35/1597-1 FUGG) as well as Center for Information Services and High Performance Computing (ZIH) at TU Dresden. 

{\small
\bibliographystyle{ieee_fullname}
\bibliography{egbib}
}

\clearpage
\pagenumbering{arabic}
\pagestyle{plain}
\setcounter{page}{1}
\appendix

\begin{center}
	\section*{Cycle-consistent Multi-graph Matching for  Self-supervised Annotation of C.~Elegans \\ \mbox{} \\ Supplementary Material}
\end{center}

\titlecontents{section}
[0.0em] 
{\vspace{.5\baselineskip}} 
{\thecontentslabel\protect\hspace{1em}} 
{} 
{\titlerule*[0.5pc]{.}\contentspage} 

\counterwithin{figure}{section}
\counterwithin{table}{section}
\counterwithin{algorithm}{section}

\section*{Table of Contents}
\startcontents[appendix-toc]
\printcontents[appendix-toc]{}{0}[1]{}

\FloatBarrier

\section{Runtimes}\label{suppl:runtimes}
All runtime experiments were conducted on a machine equipped with an AMD Ryzen Threadripper PRO 5975WX CPU (32 cores, 3.60 GHz) and 256 GB of RAM. For comparability and consistency, all experiments were performed using single-threaded processing without parallel jobs.

We present runtimes for all steps involved in our proposed unsupervised MGM pipeline in Tab.~\ref{tab:runtime1}. The alignment step encompasses all 100 training worms and 100 test worms. The Bayesian optimization parameter learning was performed on a subset of $N_{\text{learn}} = 15$ training worms. Additionally, we report the runtimes for the three different MGM solver modes (see \cref{sec:mgm}), when applied to solve the worm multi-matching problem over all 100 training worms. Finally, the evaluation runtimes include atlas construction, matching the atlas to each of the 100 test worms, and performing the final evaluation.

\begin{table}[H]
\centering
\caption{Runtimes of the unsupervised MGM pipeline.}
\begin{tabular}{lr}
\toprule
\textbf{Process} & \textbf{Duration (h)} \\
\midrule
Alignment (\cref{sec:alignment})& 8 \\
Parameter learning (\cref{sec:BO-learning}) & 81 \\
MGM Solver (\textit{Direct}) & 2 \\
MGM Solver (Sync. sparse) & 10 \\
MGM Solver (Sync. dense) & 10 \\
Atlas building \& evaluation & 0.3 \\
\bottomrule
\end{tabular}
\label{tab:runtime1}
\end{table}

We further detail the runtime of the MGM solver modes in \cref{tab:runtime2}.
While synchronization-based methods solve the MGM problem in under an hour, they require a preprocessing step involving the solution of $100 \choose 2$$ = 4950$ pairwise GM problems, which constitutes the most time-consuming part of the process. 

\begin{table}[H]
	\centering
	\caption{Runtimes of the three MGM solver modes when solving the worm multi-matching problem over all 100 training worms. Durations are given in h:min.}
	\begin{tabular}{lccc}
		\toprule
		& \textbf{Direct} & \makecell{\textbf{Sync.}\\\textbf{sparse}} & \makecell{\textbf{Sync.}\\ \textbf{dense}} \\
		\midrule
		Pairwise GM & -- & 09:36 & 09:36 \\
		MGM Solving & 02:11 & 00:04 & 00:46 \\
		\midrule
		Total & 02:11 & 09:40 & 10:22 \\ 
		\bottomrule
	\end{tabular}
	\label{tab:runtime2}
\end{table}

In contrast, the \textit{direct} mode does not require this costly initial step. Consequently, the \textit{direct} solver is overall nearly five times faster than the synchronization-based modes. However, solving the pairwise problems in parallel could in theory significantly reduce their runtime.

We attribute the difference between the \emph{sparse} and \emph{dense} synchronization mode mainly to an inefficient implementation. 
The software was written to solve sparse problems in particular and currently does not handle the many zero cost assignments of \cref{equ:synchronization-cost} well. Infinite cost assignments of sparse problems are only modelled implicitly by not storing a cost value for these assignments. This is highly beneficial to make sparse problems tractable, however, many optimizations that can be assumed for dense problems are not applied.

\section{Cost learning}
We provide details of \cref{sec:BO-learning} where the covariance and problem size parameters for our unsupervised MGM pipeline are learned based on a subset of $N_\text{learn} = 15$ training worms.


\paragraph{Learned parameters.} The model size parameters learned from $N_\text{learn} = 15$ training worms of the \texttt{200worms} dataset are given as follows:
\begin{equation}
K_\text{min} = 14, \quad \tau_\text{cen} = 1.95, \quad \tau_\text{rad} = 0.15
\end{equation}

The learned covariances are given by
\begin{equation}
\Sigma^\text{cen} =
\begin{pmatrix}
130 & 0 & 0 \\
0 & 43 & 0 \\
0 & 0 & 28
\end{pmatrix}, \quad
\Sigma^\text{rad} =
\begin{pmatrix}
156 & 0 & 0 \\
0 & 200 & 0 \\
0 & 0 & 174
\end{pmatrix}
\end{equation}
and
\begin{equation}
\quad
\Sigma^\text{off} =
\begin{pmatrix}
172 & 0 & 0 \\
0 & 32 & 0 \\
0 & 0 & 44
\end{pmatrix}
\end{equation}

\paragraph{Unassignment cost constant.} To avoid trivial empty matchings, which would result in a minimal cycle loss of zero but would fail to produce meaningful solutions, the  \emph{unassignment cost} $c_0$ is crucial. 
This constant forces the algorithm to establish correspondences by preventing an empty solution. 
We set $c_0 = 10000$ for cost learning and atlas matching, to enforce as many matchings as possible. 
For the MGM problem after cost learning, we chose $c_0 = 40$, which results, on average, in one unmatched nucleus per GM problem. 
By allowing this, we found the resulting atlas to yield slightly better results. 
However, the best value is very much dependent on the final magnitude of all the cost parameters.
As such, we recommend fine-tuning $c_0$ via a grid search or another Bayesian optimization step on a validation dataset.

\section{Additional ablation experiments}
In addition to the experiments shown in \cref{table:ablation_model}, we performed ablations on our final model without retraining.
The aim is to investigate the contribution of the second re-alignment step and the impact of individual cost terms on the final accuracy. The results are presented in \cref{table:suppl_ablation}.

\begin{table}
   \centering
\caption{Evaluating ablations on the final model without relearning. 
Keeping learned covariances constant, we investigate the impact on the final pre-atlas and atlas accuracy when linear \emph{center} ($\lambda^{\text{cen}}$), \emph{radii} ($\lambda^{\text{rad}}$) or quadratic \emph{offset vector} ($\lambda^{\text{off}}$) costs are discarded.
Notably, including radii costs leads to no improvement in the linear model. This can be explained by numerous manually curated cell segments in our training and test data, with a constant cell-independent segment radius. Thus the latter has little to no influence on the model.
In addition, shown by the \emph{single re-alignment} ablation, we tested the effect of the second re-alignment step on the training worms, which is performed after cost learning. 
Skipping it, leads to only slightly worse results.
}
   \begin{tabular}{@{}lcc@{}}
     \toprule
     Ablation & Pre-Atlas Acc. & Atlas Acc. \\
     \midrule
      $\lambda^\text{cen} = 1,\; \lambda^\text{\{rad, off\}} = 0$ & $0.953 \pm 0.024 $ & $ 0.960 \pm 0.021$\\
      $\lambda^\text{rad} = 1,\; \lambda^\text{\{cen, off\}} = 0$ & $0.260 \pm 0.189 $ & $ 0.060 \pm 0.011$\\
      $\lambda^\text{off} = 0,\; \lambda^\text{\{cen, rad\}} = 1$ & $0.953 \pm 0.023 $ & $ 0.960 \pm 0.021$\\
      $\lambda^\text{off} = 1,\; \lambda^\text{\{cen, rad\}} = 0$ & $0.967 \pm 0.016 $ & $ 0.958 \pm 0.021$\\
      Single re-alignment & $0.963 \pm 0.018 $ & $ 0.971 \pm 0.015$\\
      \midrule
      Full pipeline & $0.966 \pm 0.016$ & $ 0.972 \pm 0.015$\\
     \bottomrule
   \end{tabular}
   \label{table:suppl_ablation}
\end{table}

\section{Atlas accuracy distribution over test worms.}
We illustrate in \cref{fig:supervised_unsupervised_density} the specific accuracy value of every test worm for both our supervised and unsupervised atlas.

\begin{figure}
    \centering
    \scriptsize
    \input{figures/plot_supervised_unsupervised.pgf}
    \caption{
    Distribution and density of accuracies on the 100 test worms for the supervised and unsupervised atlas.
    Each black dot represents a distinct worm of the test set and its corresponding accuracy.
    The violin plot behind the black dots shows the approximate shape of the probability density function over all samples.
    \label{fig:supervised_unsupervised_density}}
\end{figure}

\section{Supervised atlas construction}
\label{sec:supervised_atlas_construction}

We present our construction of the supervised baseline, which is based on the approach outlined in~\cite{golland_active_2014}. Since we have access to the ground truth labels from which we can extract additional information, the supervised approach differs from the unsupervised one in the following aspects:

\paragraph{Alignment.} Similar to the unsupervised case, we first pre-process all worms as described in \cref{sec:alignment}, then
align all worms to an optimal base worm $W$. In the supervised case we follow the approach from~\cite{li_full-body_2024}. We start by fixing each training worm $W_i$ as a target. We then align the nuclei centroids of all remaining worms to $W$ using a linear transform minimizing the total Euclidean distance of corresponding centroids. From all worms aligned to $W_i$ we build a supervised atlas based on $W_i$. This results in 100 supervised atlases. Now, for each such atlas we compute the average normalized centroid distance to each of the remaining 99~training worms, and average this distance across these. We choose as our supervised baseline atlas the one atlas minimizing the average distance, and finally align all other training worms again by minimizing the total Euclidean distance of corresponding centroids. This ensures that the aligned worms fit as closely as possible to the selected base worm $W$.
Lastly, for the alignment of the test worms, we apply the average affine transformation to each worm in the test set.


\paragraph{Graph matching (GM) hyperparameters.} We adopt the sparsity parameters for the atlas-to-worm GM problems from~\cite{golland_active_2014}. 
These are given by $\tau_\text{cen} = 8$, $\tau_\text{rad} = 12$ and $K = 6$. However, our experiments suggest that the cost weight parameters $\lambda^\text{cen}, \lambda^\text{rad}, \lambda^\text{off}$ reported in~\cite{golland_active_2014} are not optimal. 
We thus finetune the hyperparameters $\lambda^\text{cen}, \lambda^\text{rad}, \lambda^\text{off} \geq 0$ using Bayesian Optimization~\cite{Ozaki_Multi_TPE_2020} where our goal is to maximize the average matching accuracy of the baseline atlas w.r.t. to all 100 training worms. 
The optimal parameters we found after 500 optimization steps are given by

\begin{equation}
\lambda^\text{cen} = 0.48, \quad \lambda^\text{rad} = 0.34, \quad \lambda^\text{off} = 0.81
\end{equation}

\paragraph{Improved supervised baseline}
We decide to use our supervised atlas as a baseline for comparison instead of those provided in~\cite{golland_active_2014} and~\cite{li_full-body_2024}, since it achieves overall higher accuracy values:

\begin{itemize}
    \item~\cite{li_full-body_2024} apply their methods on the same \texttt{200worms} dataset we used for our experiments. We achieve an average accuracy $0.96 \pm 0.02$ compared to their $0.93 \pm 0.11$. Note that we used the reported numbers of their single optimal template (i.e. atlas) experiment for comparison because this experiment corresponds to our supervised atlas building pipeline. While~\cite{li_full-body_2024} achieve higher accuracies in other atlas-to-worm matching methods ($0.95$ using “5-template before EPC” and $0.97$ using “5-template after EPC”) those methods do not directly compare, as they rely on evaluating the final accuracy based on the best five atlases instead of just one, or on an Error Prediction and Correction (EPC) approach which involves manual curation (after matching) of nuclei flagged as having a low confidence score.
    
    \item~\cite{golland_active_2014} apply their methods on a dataset consisting of 30 manually segmented and only partially ground truth annotated worms. We achieve an average accuracy $0.91 \pm 0.05$ compared to their $0.90 \pm 0.07$. Note that accuracies for this dataset were obtained by averaging the atlas accuracies in a leave-one-out fashion, i.e. we created 30 atlases where each one of the 30 worms was fixed as test target and the remaining 29 worms were used to build an atlas. The reported accuracies are the accuracies averaged across these 30 atlases.
\end{itemize}

\section{Atlas from StarDist segmentations}
\label{suppl:stardist}
\begin{figure}
    \centering
    \scriptsize
    \includegraphics[width=\columnwidth]{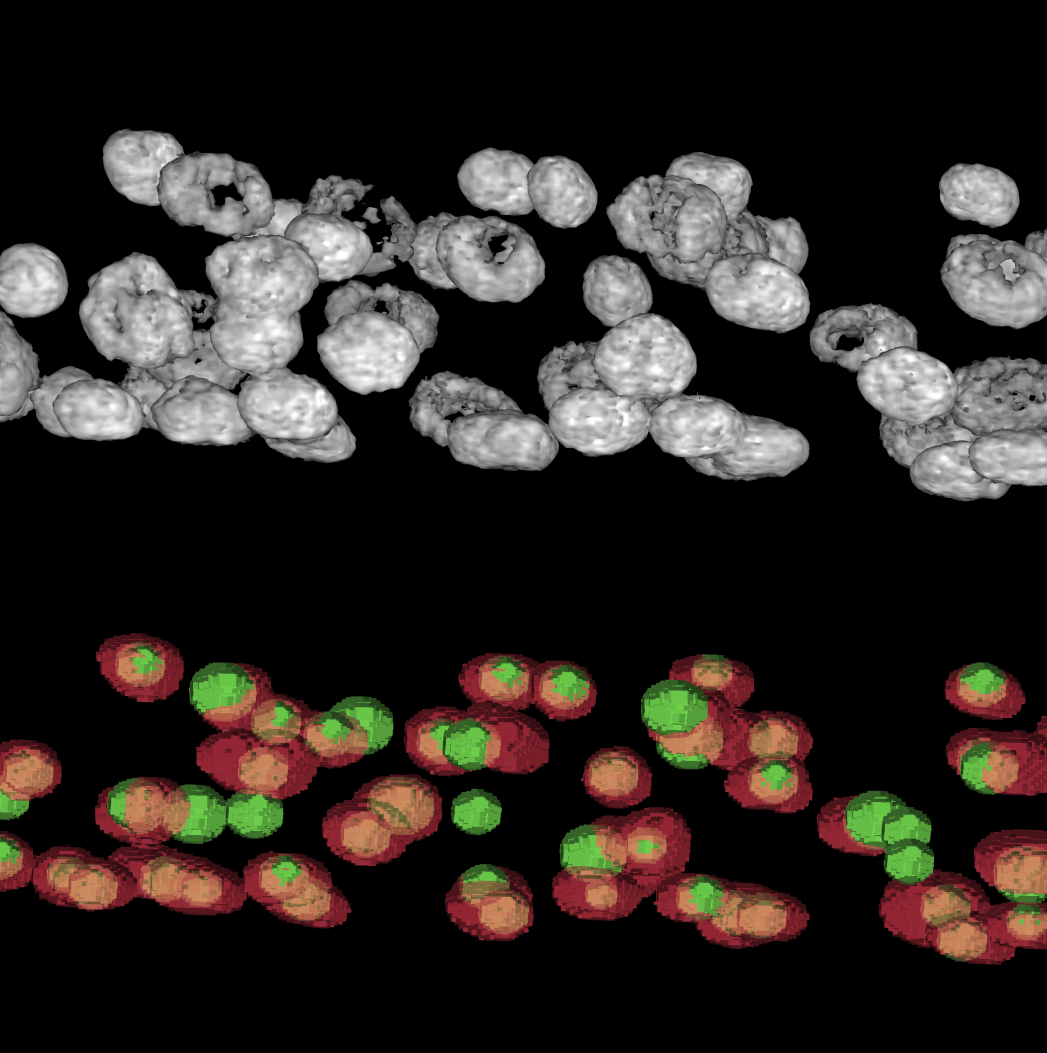}
    \caption{Excerpt of a raw 3D worm image (top) compared to the two segmentations (bottom): The manual segmentation of dataset~\cite{li_full-body_2024}, showing ground-truth annotations (green) and a segmentation predicted by \emph{StarDist} (red). The manual labels frequently appear as small spherical segments, reflecting the incomplete nature of the manual annotation process, in which nuclei were only marked coarsely rather than fully delineated.
    \label{fig:stardist_segmentation}
    }
\end{figure}

The experiments in \cref{sec:ablations-and-results} use the segmentations of the dataset~\cite{li_full-body_2024}, which are expert-corrected for segmentation errors%
\footnote{As visible in \cref{fig:stardist_segmentation}, all manually placed cell segments of dataset~\cite{li_full-body_2024} are perfect spheres. 
Here, \emph{expert-correction} involved only removing incorrect segments and accurately placing nuclei center points. 
Consequently, these annotations do not constitute a true ground truth \wrt the segmentations themselves. 
Nevertheless, we will continue to refer to this data as the ground truth.}. 
In the following, we investigate the impact of using automatically generated segmentations.

\paragraph{Generating automatic segmentations.}
We apply \emph{StarDist}~\cite{startdist_weigert2022,weigert2020stardist-3D}, a state-of the art cell instance segmentation model, to our training and test images.
For 3D cell segmentation, \emph{StarDist} provides only a \emph{demo model}, which we used without retraining. 
Since we assess sensitivity to imperfect segmentations, using a non-optimized model is acceptable.
\cref{fig:stardist_segmentation} visually compares the given ground truth segmentations against the segmentations returned by \emph{StarDist}.

On average, $528 \pm 23$ nuclei were segmented, which is notably less than the $ 558 $ that are present in each imaged worm.
The intersection over union (IoU) between the ground truth and the \emph{StarDist} segmentations is $0.40 \pm 0.03$, which is low but expected due to the rough spherical shape approximations in the ground truth and the lower number of cells segmented by \emph{StarDist}.

\paragraph{Transferring ground truth labels.}

To evaluate our unsupervised and supervised approaches with the \emph{StarDist} segmentations, nuclei annotated with labels are required.
We transfer these labels from the ground truth segmentation based on cell overlap: 
Each generated cell segment is matched to the ground truth segment with which it shares the largest voxel intersection.
To ensure consistency, we apply the same heuristic we used to assign ground truth labels to an unsupervised atlas (see
\cref{sec:unsupervised-atlas-labels}), guaranteeing that each ground truth nucleus is assigned to at most one \emph{StarDist} segment.

The resulting matching yields a labeling of the predicted nuclei by inheriting the labels from the corresponding ground truth nuclei. 
Using this approach, we annotate on average $503 \pm 19$ nuclei per worm in the \emph{StarDist} segmentations, leaving approximately $25$ nuclei per worm unlabeled.

Despite this, both the supervised and unsupervised atlases are able to recover around $545$ distinct nuclei, due to aggregation over all worms during atlas construction.

\paragraph{Cost learning. } For the \emph{StarDist} segmentation in the unsupervised setting
we have obtained the following parameters:
\begin{equation}\label{equ:star-dist-unsupervised-atlas-params-1}
K_\text{min} = 16, \quad \tau_\text{cen} = 9.08, \quad \tau_\text{rad} = 7.30
\end{equation}
\begin{equation}\label{equ:star-dist-unsupervised-atlas-params-2}
\Sigma^\text{cen} =
\begin{pmatrix}
176 & 0 & 0 \\
0 & 50 & 0 \\
0 & 0 & 24
\end{pmatrix}, \quad
\Sigma^\text{rad} =
\begin{pmatrix}
65 & 0 & 0 \\
0 & 188 & 0 \\
0 & 0 & 10
\end{pmatrix}
\end{equation}
and
\begin{equation}\label{equ:star-dist-unsupervised-atlas-params-3}
\quad
\Sigma^\text{off} =
\begin{pmatrix}
200 & 0 & 0 \\
0 & 181 & 0 \\
0 & 0 & 182
\end{pmatrix}
\end{equation}

In the supervised setting, we obtained the following weights:

\begin{equation}\label{equ:star-dist-supervised-atlas-params}
\lambda^\text{cen} = 0.21, \quad \lambda^\text{rad} = 1.63, \quad \lambda^\text{off} = 1.35
\end{equation}

\subsection{Accuracy evaluation (\cref{table:results_stardist}).}
As shown in \cref{table:results_stardist}, the accuracy gap between the supervised and unsupervised models increases substantially when using the \emph{StarDist} segmentation, compared to the ground-truth segmentation results in \cref{table:supervised_comparison}. Notably, the performance gap between the \emph{Unlearned quadratic} model and the \emph{Full unsupervised pipeline} also widens considerably, clearly demonstrating the benefit of our learning approach and its robustness to imperfect segmentations.

That said, we found the \emph{Pre-processing} step (see \cref{sec:alignment}) to be the most vulnerable component to segmentation errors in the \emph{StarDist} output, and a major contributor to the observed accuracy drop. While the pre-processing aligned worms within $1$--$5^\circ$ using ground-truth segmentations, the alignment error increased to nearly $40^\circ$ with \emph{StarDist} segmentations; see \cref{fig:error_alignment_angle}. Such large misalignment often could not be fully corrected by the subsequent steps based on pairwise GM described in the \emph{Alignments} paragraph of \cref{sec:alignment}.

The supervised pipeline was considerably more robust to these pre-processing issues, as it uses ground-truth correspondences for alignment during training (see \cref{sec:supervised_atlas_construction}). The resulting atlas parameters~\eqref{equ:star-dist-supervised-atlas-params} place the strongest weight on shape and offset features, which are largely insensitive to misalignment, yielding better test accuracy. In contrast, the unsupervised parameters~\eqref{equ:star-dist-unsupervised-atlas-params-2}--\eqref{equ:star-dist-unsupervised-atlas-params-3} assign significantly higher weight to absolute coordinates, making them substantially more sensitive to alignment errors and therefore more prone to mismatches during GM.

\paragraph{Additional ablation experiment (\cref{table:results_stardist}).}
The variant \emph{Learned} $\lambda^{\text{\{cen, rad, off\}}}$ extends the \emph{unlearned quadratic} model by performing 500 Bayesian optimization iterations to estimate the weights $\lambda^{\text{\{cen, rad, off\}}}$:

\begin{equation}
\lambda^\text{cen} = 0.03, \quad \lambda^\text{rad} = 1.55, \quad \lambda^\text{off} = 0.06
.
\end{equation}

Analogously to the supervised setting, this model is less sensitive to pre-processing errors, primarily due to the higher weight of the shape term $\lambda^{\text{rad}}$. 
Due to that, this simplified model even outperforms our \emph{Full unsupervised pipeline}.

\paragraph{Details on the pre-processing errors with StarDist segmentation (\cref{fig:error_alignment_angle}).}
Due to the high cell density in the worm brain, the head--tail alignment performed equally well for the \emph{StarDist} and the ground-truth segmentations. However, the left--right (dorsal--ventral) alignment was substantially less accurate for \emph{StarDist}, which we attribute to its segmentations being noticeably less symmetric. We validated this by manually re-rotating the test worm with the largest accuracy drop about its head--tail axis, which immediately improved its accuracy from \(5\%\) to \(64\%\).

\begin{table*}
   \centering
\caption{Accuracy based on the \emph{StarDist} segmentation for different atlas construction methods. 
Unsupervised and supervised pipelines constructed and evaluated as in \cref{sec:results_comparison}.
}
   \begin{tabular}{@{}lcc@{}}
     \toprule
        Method  
        & \makecell{Pre-Atlas Accuracy\\(Training data)} 
        & \makecell{Atlas Accuracy\\(Test data)} \\
     \midrule
      Unlearned quadratic & $0.668 \pm 0.199 $ & $ 0.609 \pm 0.192 $\\
      Learned $\lambda^{\text{\{cen, rad, off\}}}$ & $\boldsymbol{0.858 \pm 0.182} $ & $ 0.829 \pm 0.062 $\\
      \midrule
      Full unsupervised pipeline & $0.827 \pm 0.196$ & $ 0.809 \pm 0.147 $\\
      Supervised pipeline & \textemdash{} & $\boldsymbol{0.916 \pm 0.029} $\\
     \bottomrule
   \end{tabular}
   \label{table:results_stardist}
\end{table*}

\begin{figure}
    \centering
    \scriptsize
    \input{figures/accuracy_loss_vs_alignment_error.pgf}
    \caption{
    Relationship between alignment error and accuracy loss. Each point corresponds to one test worm. During pre-processing (\cref{sec:alignment}), worms are rotated about the head–tail axis to maximize left–right symmetry. The x-axis shows the difference between the rotation angles obtained from the ground-truth and the StarDist segmentations. The y-axis shows the resulting drop in matching accuracy when replacing ground-truth segmentation with StarDist, using the respective unsupervised atlases.
    \label{fig:error_alignment_angle}
    }
\end{figure}

\end{document}